\documentclass{article}
\usepackage{spconf,amsmath,graphicx}
\usepackage{float}
\usepackage{hyperref}

\usepackage [english]{babel}
\usepackage [autostyle, english = american]{csquotes}
\MakeOuterQuote{"}

\title{Fast, Self Supervised, Fully Convolutional Color Normalization \\ of H\&E Stained Images}

\name{Abhijeet Patil$^{*1}$\qquad Mohd. Talha$^{*1}$\thanks{$*$ These authors contributed equally.}\qquad Aniket Bhatia$^{1}$\qquad Nikhil Cherian Kurian$^{1}$ \vspace{-0.48cm}}
\address{\textit{Sammed Mangale$^{1}$\qquad Sunil Patel$^{2}$ \qquad Amit Sethi$^{1}$}\\\\
$^{1}$Department of Electrical Engineering, IIT Bombay, India,  $^{2}$ Nvidia, Mumbai, India}

\begin{document}

\maketitle

\begin{abstract}
Performance of deep learning algorithms decreases drastically if the data distributions of the training and testing sets are different. Due to variations in staining protocols, reagent brands, and habits of technicians, color variation in digital histopathology images is quite common. Color variation causes problems for the deployment of deep learning-based solutions for automatic diagnosis system in histopathology. Previously proposed color normalization methods consider a small patch as a reference for normalization, which creates artifacts on out-of-distribution source images. These methods are also slow as most of the computation is performed on CPUs instead of the GPUs. We propose a color normalization technique, which is fast during its self-supervised training as well as inference. Our method is based on a lightweight fully-convolutional neural network and can be easily attached to a deep learning-based pipeline as a pre-processing block. For classification and segmentation tasks on CAMELYON17 and MoNuSeg datasets respectively, the proposed method is faster and gives a greater increase in accuracy than the state of the art methods.
\end{abstract}
\vspace{-5pt}
\vspace{-5pt}

\section{Introduction}
\vspace{-5pt}
\label{sec:intro}
Deep learning-based algorithms have been used for a variety of diagnostic tasks for histopathology images of tissue samples stained using hematoxylin and eosin (H\&E)~\cite{golatkar,gleason,nature_lung}. Most of these algorithms yield expected results when training and testing data have a similar color appearance. However, the performance of these algorithms drops drastically when tested with images from other labs~\cite{sethi2016empirical}, primarily due to variations in stain appearance that result from differences in scanner sensor responses, H\&E staining protocols, reagents, and habits of technicians. Such variations in colors are not an issue for expert pathologists, as they are trained to prioritize morphological features over strict color definitions of H\&E stained slides. However, stain colors of training data create bias for a deep learning algorithm, which can cause problems at the time of deployment.


\begin{figure*}[h]
\centering
\includegraphics[width=14cm]{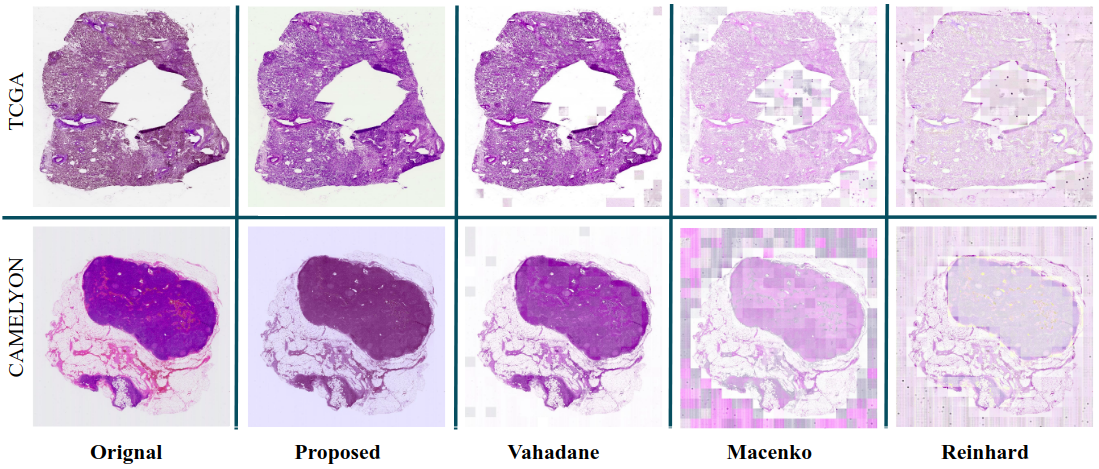}
\caption{Visual comparison between the proposed ColorNormNet and other popular color normalization methods on representative WSIs from TCGA (top) and CAMELYON (bottom) datasets. Notice the image artifacts introduced by the other techniques.}
\label{VisualComparison}
\end{figure*}

Color normalization is used to transform input samples in computational pathology to a pre-defined color space to aid deep learning algorithms~\cite{sethi2016empirical,khan,reinhard,macenko,vahadhane,gautam}. Most of the practical color normalization approaches depend on matrix factorization for color deconvolution. Such approaches take the reference color matrix from a small patch of the H\&E stained image, which causes problems when the source image is from significantly different tissue region (anatomically speaking) than the target patch~\cite{gautam}. Additionally, in order to estimate a robust factorization for the entire whole slide image (WSI), such methods often become computationally expensive. On the other hand, more recent color normalization techniques have turned to deep learning, particularly generative adversarial networks~\cite{isbigan,staingan,miccai19,inceptionlab}, although such methods are computationally quite expensive to train, and cannot be easily integrated with downstream deep learning pipelines.

Our innovation lies in departing from matrix factorization as well as GANs. Instead, we use self-supervised learning, where a lightweight neural network is trained to estimate the color shift needed in each channel to match a pre-determined target stain in appearance. Compared to the state-of-the-art color normalization algorithms, the advantages of our algorithm are that it:
\begin{itemize}
\setlength\itemsep{.1em}
\item Trains fast (Section~\ref{sec:proposed}),
\item Tests faster (Table~\ref{TestTime}), 
\item Has a more positive impact on the accuracy of downstream tasks, such as classification on CAMELYON17 dataset~\cite{camelyon}  and segmentation on MoNuSeg dataset~\cite{monuseg} (Table~\ref{Accuracy}), 
\item Has fewer post-normalization image artifacts (Figure~\ref{VisualComparison}), and
\item Integrates easily with deep learning pipelines (Figure~\ref{BlockDiagram}).
\end{itemize}

\section{Related Work}
Most of the color normalization methods compute a stain color matrix $W \in \rm I\!R ^{3\times2}$ and stain density maps $H$ of source and target images and try to project the color matrix of the source image to that of the target image.  Matrix factorization is a popular technique used for color deconvolution. Mecenko et al.~\cite{macenko} had framed color deconvolution as singular value decomposition (SVD) problem to guarantee the solution for an optimization problem. Reinhard et al.~\cite{reinhard} used methods to match histograms for color densities in source and target images. Vahadane et al.~\cite{vahadhane} added sparsity to non-negative matrix factorization (NMF) to achieve better results. These approaches are patch-based. That is, the reference color matrix is taken from a small patch of the H\&E stained image. This causes problems when the source image is from significantly different tissue region than the target patch~\cite{gautam}. The workaround in these methods against getting a biased estimate of the stain color matrix $W$ of source image is to sample the matrix from the entire image, which is computationally expensive. The proposed method does not compute the stain color matrix. 

Deep learning-based approaches have recently been employed for color normalization, especially those using generative adversarial networks (GANs). For example, Aicha et al.~\cite{aicha} proposed an end-to-end deep learning method to learn stain normalization along with specific diagnosis tasks, where stain normalization is learned adversarially. Farhad et al.~\cite{isbigan} and Shaban et al.~\cite{staingan} use info-GAN~\cite{infogan} and cycleGAN~\cite{cyclegan} respectively in their color normalization techniques. Niyun et al.~\cite{miccai19} uses a color matrix of the source image as auxiliary information for a generator based on cycleGAN for better learning. GANs are hard to train and require higher compute at inference time. Moreover, GANs can change microscopic morphological features which may important for diagnosis. Dwarikanath et al.~\cite{inceptionlab} tries to resolve this problem by adding auxiliary feature from CNN trained for nucleus segmentation task, while SAASN~\cite{saasn} uses self-attention to preserve the local context, which makes their training process even more complicated. Additionally, Stanosa et al.'s model~\cite{stanosa} learns to pick a target template for normalization of H\&E image according to the tissue type of the source image.

\section{Proposed Method}
\label{sec:proposed}

\begin{figure*}[h]
  \centering \includegraphics[width=16cm]{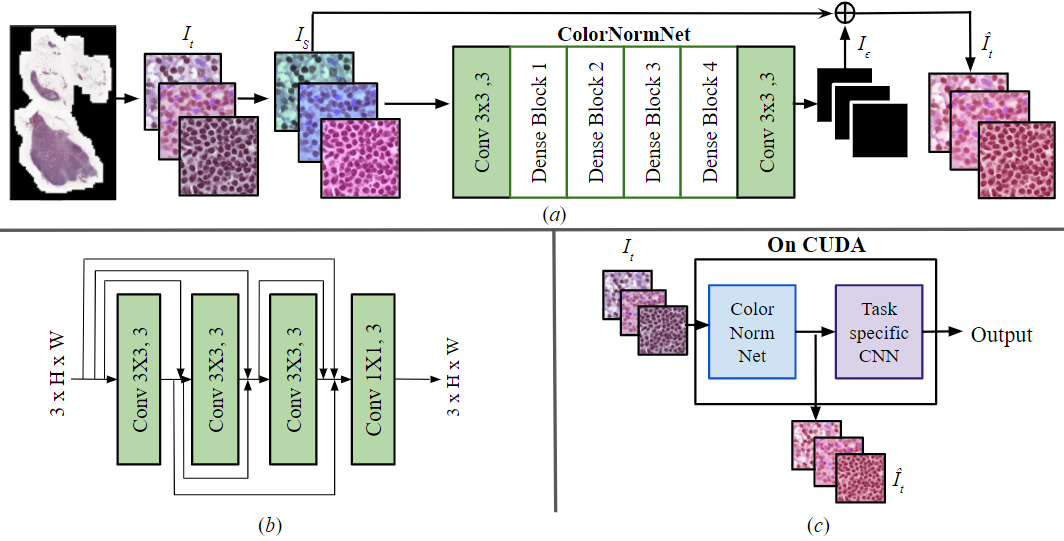}
  \caption{Working of the proposed method color normalization network (ColorNormNet), including (a) the training pipeline, (b) the architecture of Dense Block used in ColorNormNet, and (c) how the trained ColorNormNet can be attached to task-specific CNN as a pre-processing block.}
  \label{BlockDiagram}
\end{figure*}

The proposed method alleviates several problems in the existing factorization-based and deep learning-based color normalization methods based on the following features:
\begin{itemize}
\setlength\itemsep{.1em}
\item It uses a self-supervised fully convolutional neural network that is lightweight (2,352 trainable parameters) and easy to train. Training the proposed CNN architecture takes only half an hour on Nvidia 1080Ti GPU,  and inference takes about a minute and a half per gigapixel. By comparison, GAN-based color normalization model can increase the inference time drastically as they have millions of parameters. Also, for template-based methods computation is performed on CPUs which makes inference time slower.
\item In the proposed method, WSI or set of WSIs with target color appearance acts as target domain, which can include a variety of tissue types. This eliminates problems caused by template-based methods such as the introduction of artifacts.
\item We can change the target color appearance by training proposed architecture on patches from new target WSIs easily, which is difficult in GAN based color normalization methods.
\end{itemize}

The overall block diagram is shown in Figure~\ref{BlockDiagram}. We next describe the self-supervision, the architecture, and the training details.

\textbf{Self-supervision:} In self-supervision, we use label that comes for "free" with data. There are several self-supervised approaches to learn representation of unlabeled data~\cite{selfsup}. In our setting, we train a deep learning model to predict the offset which is added synthetically to an unaltered patch sampled from WSI, $W_{target}$. Target domain $D_{target} = \{I_1, I_2, ...I_N\}$ is a set of images, such that $I_n$ is the $n^{th}$ patch in $W_{target}$. We perturb the $R$ and $B$ channel of a target histopathology tissue image $I_t \in D_{target}$ by adding a random offset to it to get a synthetic source image, $I_{s}$ as shown in following equation.
\begin{equation}
\begin{aligned}
I_{s}^{R} = I_{t}^{R} + \epsilon_1 ;
I_{s}^{B} = I_{t}^{B} + \epsilon_2 ;
I_{s}^{G} = I_{t}^{G}
\end{aligned}
\end{equation}
where $\epsilon_1$ and $\epsilon_2$ are random numbers drawn independently from uniform distribution $Unif(-0.2,0.2)$. We do not add offset to $G$ channel of $I_t$ because stain variation generally do not effect intensity of $G$ channel of image. Figure \ref{BlockDiagram} shows sample target image and synthetically generated source image corresponding to it. 

\textbf{Model Architecture:} Our model uses a modified version of DenseNet blocks. We have used four DenseNet blocks with only three convolutional filters at each layer. We hypothesize that learning the color offset $I_\epsilon$ does not require to learn too many features as long as the prediction is obtained using a large enough receptive field. To increase the receptive field, we have used dilated convolutional layers in Dense Block 2 and Dense Block 3. Each convolutional layer in Dense Block is followed by leaky ReLU and a batch normalization layer. We have not used any pooling operation in our architecture. The size of feature map remains equal to size of the input image throughout the model. As we have used only three filters in every convolutional layer, the number of trainable parameters is also very small. Our model, which we call \emph{ColorNormNet}, contains only 2,352 trainable parameters.

\textbf{Training details:} We optimize parameters of a DenseNet ~\cite{densenet} inspired fully convolutional neural network $f(I_s; \theta_f)$ to predict offset tensor $I_\epsilon$ such that $\hat{I}_t = I_s + I_\epsilon$. We train ColorNormNet by optimizing following loss function
\begin{equation*}
    L(I_t, \hat{I}_t) = ||\hat{I}_t-I_t||_1 + \lambda||\hat{I}_t-I_t||_2^2
\end{equation*}
where $\lambda$ is a hyper-parameter which balances $L1$ and $L2$ loss terms. We used $\lambda=0.1$ for training ColorNormNet. We train this model with the patches from a target WSI. We observed that even $1000$ patches are sufficient to train a color normalization network. Training the proposed CNN architecture takes about half an hour (800-1000 iterations) to converge on a Nvidia 1080Ti GPU with a batch size of 128 and patches of size $256\times256$, and Adam optimizer with learning rate 0.001. Figure \ref{BlockDiagram} shows training and inference pipeline along with architecture of CNN model. We can also train a model with arbitrary patch size as the proposed model is fully convolutional.

\begin{table*}[h]
\centering
\caption{Comparison of classification and segmentation results for different color normalization methods.}
\label{Accuracy}
\begin{tabular}{c|cccccc|ccccc}
                   & \multicolumn{6}{c|}{Classification Results (AUC)}              & \multicolumn{5}{c}{Segmentation Results} \\ \hline
\textbf{Method}    & Center 1 & Center 2 & Center 3 & Center 4 & Center 5 & Average & Dice   & AJI  & PQ    \\ \hline
W/O Norm.  & 0.83    & 0.83    & 0.67    & 0.87    & 0.80    & 0.80   & 0.78  & 0.52   & 0.55 \\
Reinhard   & 0.83    & 0.82    & 0.80    & 0.87    & 0.81    & 0.83   & \textbf{0.81}  & \textbf{0.53}  & 0.54   \\
Macenko   & 0.77    & 0.83    & 0.84    & 0.86    & 0.76    & 0.81   & 0.66  & 0.39  & 0.46   \\
Vahadane & 0.81    & 0.82    & \textbf{0.86}    & 0.86    & 0.88    & 0.84   & 0.74  & 0.40  & 0.49   \\
ColorNormNet   & \textbf{0.90}    & \textbf{0.99}    & 0.79    & \textbf{0.98}    & \textbf{0.93}    & \textbf{0.91}   & 0.80  & \textbf{0.53}  & \textbf{0.56}  
\end{tabular}
\end{table*}
\begin{table}[h]
\centering
\caption{Comparison of computation times for different color normalization methods to normalize WSIs of size one giga-pixel}
\label{TestTime}
\begin{tabular}{|c|c|}
\hline
Method             & Time (seconds) \\ \hline
Vahadane  & 2642           \\ \hline
Reinhard   & 260           \\ \hline
Mecenko    & 2139           \\ \hline
Goutham-CPU & 360           \\ \hline
Goutham-GPU & 120           \\ \hline
ColorNormNet-CPU (our)           & 485           \\ \hline
ColorNormNet-GPU (our) & \textbf{87}           \\ \hline
\end{tabular}
\end{table}

\section{Experiments and Results}
We trained ColorNormNet on patches extracted from a randomly selected WSIs from the CAMELYON17 dataset~\cite{camelyon}. We compared the inference times for ColorNormNet with four state of the art methods -- by Vahadane et al.~\cite{vahadhane}, Reinhard et al.~\cite{reinhard}, Macenko et al.~\cite{macenko}, and Ramakrishna et al.~\cite{gautam}. We extracted the patches from WSI with more than one giga-pixel in size and observed the inference time for each method. Not only is ColorNormNet faster than the other methods, but it also does not produce any artifacts, as it estimates a single offset for the entire image. ColorNormNet also generalizes on the dataset from other centers and different organs. We have normalized WSIs from the TCGA dataset and observed that ColorNormNet yields better normalization on it. Figure~\ref{VisualComparison} shows color normalization results on a sample WSI each from CAMELYON17 lymph node and TCGA lung adenocarcinoma datasets. A comparison of test time per giga pixel for different techniques, including ours, is shown in Table~\ref{TestTime}.

We used CAMELYON17~\cite{camelyon} and MoNuSeg~\cite{monuseg} to evaluate ColorNormNet. CAMELYON17 contains a total of 500 WSIs, out of which 400 WSIs are of normal tissue while remaining 100 WSIs contain tumor region within them. This dataset is well-suited for the evaluation since the data is taken from five different centers and there is a significant staining variation amongst them. Out of 100 positive WSIs, only 50 WSIs (10 WSIs from each center) are annotated with the tumor regions. We used these 50 positive WSIs in our training and testing data along with the available negative WSIs. We split data from each center into training, validation, and testing with about 50\%, 30\%, and 20\% slides in each split respectively. We got following train-validation-test split for the centers 1: 33(4)-22(4)-19(2), center 2: 33(4)-20(3)-15(3), center 3: 41(5)-24(2)-19(3), center 4: 31(4)-19(3)-20(3), and center 5: 32(3)-20(4)-19(3) (Numbers in parentheses indicate the number of positive WSIs in split). We trained a ResNet50~\cite{resnet} architecture for the classification task. We used batch size of 32 with a learning rate of 0.001 and weight decay of 0.01 for the training. The classification model was trained and validated with data from one center and tested with test data from each center separately. We report the average of AUC obtained on test data of five data centers in Table~\ref{Accuracy}. ColorNormNet achieves better average AUC across the centers by a large margin.

We also evaluated ColorNormNet on nucleus segmentation task. We used MoNuSeg~\cite{monuseg} for evaluation. MoNuSeg dataset contains data from nine different organs with structural and stain variation across different images. It contains a total of 30 annotated images. We split these images in training, validation and testing datasets with 20, 5 and 5 images in each split respectively. We trained HoVer-Net~\cite{hovernet} five times using the normalized images with different methods and averaged the Dice score, aggregated Jaccard index (AJI), and panoptic quality (PQ) across the five runs for each normalization method. Table~\ref{Accuracy} shows that ColorNormNet gives better AJI and PQ than other normalization methods.
\vspace{-5pt}
\vspace{-5pt}
\vspace{-5pt}
\section{Conclusion}
\vspace{-5pt}
Color normalization is an often-used image pre-processing method for deep learning based computational pathology tasks. We have proposed a color normalization method based on a fully convolutional neural network, which is trained in a self-supervised manner. The proposed ColorNormNet is computationally fast during both training and inference, and it yields a neural network that can be attached as pre-processing block to task specific CNNs. We have also validated ColorNormNet on tumor classification and nucleus segmentation task and shown that the accuracy of these downstream tasks improve more using ColorNormNet as compared to other color normalization methods.

\vspace{-5pt}

\section{Compliance with Ethical Standards}
\vspace{-5pt}
\label{sec:ethics}
This research study was conducted retrospectively using human subject data made available in open access by \cite{camelyon, monuseg}. Ethical approval was not required as confirmed by the license attached with the open access data.
\vspace{-5pt}
\section{Acknowledgments}
\vspace{-5pt}
No funding was received for conducting this study. The authors have no relevant financial or non-financial interests to disclose.
\vspace{-5pt}

\bibliographystyle{IEEEbib}
\bibliography{refs}

\end{document}